%% file: 0_main.tex
\pdfoutput=1

\documentclass[11pt]{article}

\usepackage{EMNLP2022}

\usepackage{times}
\usepackage{latexsym}
\usepackage{subfigure}
\usepackage{xfrac}
\usepackage{colortbl}
\usepackage{booktabs}  
\usepackage{amsfonts}  
\usepackage{nicefrac}
\usepackage{multirow}
\usepackage{amsmath}
\usepackage{tabularx}
\usepackage{makecell}
\usepackage{xcolor}
\usepackage{amssymb}
\usepackage{bbding}
\newcommand{\lb}{\mbox{$\langle$}}
\newcommand{\rb}{\mbox{$\rangle$}}
\usepackage[T1]{fontenc}
\usepackage[utf8]{inputenc}
\usepackage{microtype}
\usepackage{inconsolata}
\usepackage{xparse}

\newcommand*{\email}[1]{\texttt{#1}}

\NewDocumentCommand{\heng}
{ mO{} }{\textcolor{red}{\textsuperscript{\textit{Heng}}\textsf{\textbf{\small[#1]}}}}

 \NewDocumentCommand{\zoey}
    { mO{} }{\textcolor{orange}{\textsuperscript{\textit{Zoey}}\textsf{\textbf{\small[#1]}}}}

\title{Open-Vocabulary Argument Role Prediction for Event Extraction}

\author{
Yizhu Jiao \quad
Sha Li \quad
Yiqing Xie \quad
Ming Zhong \quad
Heng Ji \quad
Jiawei Han \quad
\\
{University of Illinois at Urbana-Champaign, IL, USA} \\
\email{\{yizhuj2, shal2, xyiqing2, mingz5, hengji, hanj\}@illinois.edu} \\
}

\begin{document}
\maketitle
\begin{abstract}
The argument role in event extraction refers to the relation between an event and an argument participating in it. 
Despite the great progress in event extraction, existing studies still depend on roles pre-defined by domain experts. 
These studies expose obvious weakness when extending to emerging event types or new domains without available roles. 
Therefore, more attention and effort needs to be devoted to automatically customizing argument roles. 
In this paper, we define this essential but under-explored task: \textbf{open-vocabulary argument role prediction}. 
The goal of this task is to infer a set of argument roles for a given event type.
We propose a novel unsupervised framework, \textsc{RolePred} for this task.
Specifically, we formulate the role prediction problem as an in-filling task and construct prompts for a pre-trained language model to generate candidate roles.
By extracting and analyzing the candidate arguments, the event-specific roles are further merged and selected.
To standardize the research of this task, we collect a new event extraction dataset from WikiPpedia including 142 customized argument roles with rich semantics. 
On this dataset, \textsc{RolePred} outperforms the existing methods by a large margin. Source code and dataset are available on our GitHub repository\footnote{\url{https://github.com/yzjiao/RolePred}}.

\end{abstract}

\input{section/1_intro}

\input{section/5_related}

\input{section/2_method}

\input{section/3_dataset}

\input{section/4_exp}

\input{section/6_conclusion}

\section*{Limitations}

\textsc{RolePred} is proposed based on the assumption that most arguments are named entities.
It mainly focuses on entity arguments in raw texts.
However, although non-entity arguments are relatively rare, they also play an important semantic part in lots of events.
Our framework may get hindered when predicting roles for such non-entity arguments. 
Therefore, our next step is broader coverage of roles for different types of arguments.

In addition, our framework takes a set of related documents as input.
It requires sufficient event instances for salient role selection.
Also, the quality of generated argument roles heavily depends on document selection.
Thus, for the given event type, retrieving representative documents of limited quantity can be considered an interesting topic for argument role prediction.

Furthermore, most of the existing work defines argument roles for an event type rather than an individual event instance. 
These argument roles are shared by multiple event instances of the same type.  
Nevertheless, different event instances can have personalized characteristics. 
For example, Magnitude is an argument role shared by all earthquakes, but Number of Landslides Caused can be a specific role to certain earthquakes.
These specific roles can assist to identify specified and important arguments for event extraction.
Accordingly, we expect to customize roles for one event instance in future work.

\section*{Acknowledgements}
We thank Muhao Chen at USC, Yunyi Zhang, Tingcong Liu, and Yu Meng at UIUC for their helpful discussions and feedback. 
We would also like to thank anonymous reviewers for valuable comments and suggestions.
Research was supported in part by US DARPA KAIROS Program No. FA8750-19-2-1004 and INCAS Program No. HR001121C0165, National Science Foundation IIS-19-56151, IIS-17-41317, and IIS 17-04532, and the Molecule Maker Lab Institute: An AI Research Institutes program supported by NSF under Award No. 2019897, and the Institute for Geospatial Understanding through an Integrative Discovery Environment (I-GUIDE) by NSF under Award No. 2118329. Any opinions, findings, and conclusions or recommendations expressed herein are those of the authors and do not necessarily represent the views, either expressed or implied, of DARPA or the U.S. Government. The views and conclusions contained in this paper are those of the authors and should not be interpreted as representing any funding agencies.

\bibliography{anthology,custom}
\bibliographystyle{acl_natbib}

\input{section/7_appendix}

\end{document}

%% file: section/1_intro.tex
\section{Introduction}

\begin{figure}[t]
    \centering
    \includegraphics[width=\linewidth]{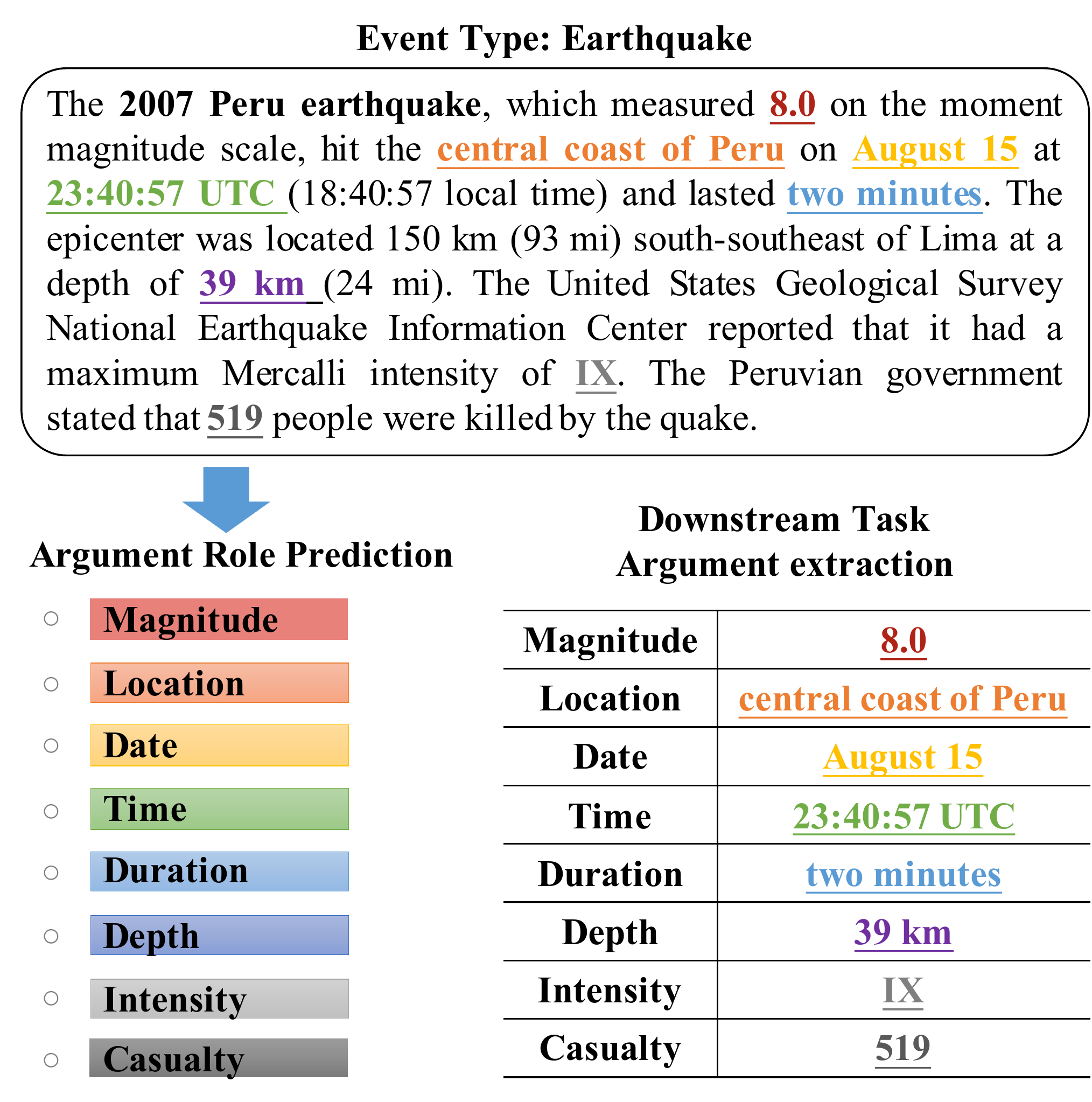}
    \caption{An example of the argument role prediction task and its downstream task.}
    \label{fig:task}
    \vspace{-4mm}
\end{figure}

Great progress has been made on event extraction in recent years, however,
most of the existing studies still rely on hand-crafted ontologies 
\cite{grishman1996message, ji2008refining, LinACL2020, du2020event, liu2020event, zhou2021role, li2021document}.
Event ontologies such as Propbank~\cite{kingsbury2003propbank} and FrameNet~\cite{baker1998berkeley} 
take years, even decades, to construct.
At the center of such ontologies lie argument roles, which capture the relation between an event and an argument participating in it. For instance, the \texttt{Transport} event type has 5 roles: \texttt{Agent}, \texttt{Artifact}, \texttt{Vehicle}, \texttt{Origin} and \texttt{Destination}.
These roles are typically specific to the event type and semantically meaningful role names can directly benefit argument extraction quality.
While human-constructed ontologies suffice for closed-domain applications, it requires extra human effort to extend to emerging event types or new domains. 
To overcome such difficulty, some studies attempt to automatically induce argument roles for given event types
\cite{huang2016liberal, yuan2018open, liu2019open}. 
These methods usually define a glossary including possible role names with general semantics, such as \texttt{Time}, \texttt{Place}, and \texttt{Value}, and then pick a subset as argument roles.
Since role names are restricted to a limited vocabulary, they do not reflect the uniqueness of event types, such as the \texttt{Magnitude} of an earthquake, or the \texttt{Host} of a ceremony. 
Hence, predicting role names from an open vocabulary is necessary for broad coverage of event semantics.

In this paper, we introduce an essential but under-explored task for event extraction: \textbf{open-vocabulary argument role prediction}. 
This task aims to infer a set of argument role names for a given event type to describe the crucial relations between the event type and its arguments.
As shown in Figure \ref{fig:task}, for the \texttt{Earthquake} event type, given some related documents, we want to output key argument role names such as \texttt{magnitude}, \texttt{intensity}, \texttt{depth}, \texttt{deaths}, and \texttt{injuries}. 
These semantically meaningful roles can be directly used in the downstream event extraction task \cite{huang2018zero, liu2020event, lyu2021zero}. 
However, this task poses new challenges: (1)
\emph{decoupling argument role prediction from argument extraction}: For event extraction, roles and arguments are closely interdependent, one of which is pivotal to determining the other, and predicting argument roles for unknown arguments is a pressing problem; and (2) 
\emph{customizing argument roles from an open vocabulary}:  To cover board domains, we need to go beyond the predefined candidate vocabulary, and, the generated roles should be personalized for each event type so that they can reflect the unique features of different event types.

To tackle these challenges, we propose a novel unsupervised framework, \textsc{RolePred}. 
Given an event type and a set of documents, \textsc{RolePred} predicts the argument roles by three  components including candidate role prediction, candidate argument extraction, and argument role selection. 
Concretely, to decouple roles from unknown arguments, we assume that named entities are more likely to be arguments.
Based on this assumption, we regard the named entities in the text as possible arguments.
Then, we predict their candidate role names by casting it as a prompt-based in-filling task \cite{raffel2020exploring}.
Note that, we allow the pre-trained model \cite{raffel2020exploring} to fill in a variable-length mask span instead of one single mask. 
Yet, those generated roles are still noisy.
Therefore, considering the inter-dependency between roles and arguments, we extract arguments with QA models for further role selection and merging.
Finally, the event-specific roles are obtained to serve for event extraction.
In this way, generated roles are sufficiently fine-grained and event-specific. 

Existing event extraction datasets have limited coverage of event types and insufficient refinement of argument roles \cite{grishman1996message, li2021document, ebner2020multi}.
Thus, to support the research in argument role prediction, we collect a new event extraction dataset from Wikipedia named RoleEE. 
In statistics, our dataset contains 50 event types and 142 argument role types, 
much more than the number of argument roles in the existing dataset (5 in MUC-4 \cite{doddington2004automatic} and 65 in RAMS \cite{ebner2020multi}).
Besides the general roles, such as date and location, there are personalized roles for each event type, such as Accelerant for a Fire event, and Magnitude for an earthquake event, which 
carry rich semantics and assist to extract detailed arguments in events. 
Besides, our dataset focuses on the extraction of the main event in each document, that is, one-event-per-document. 
This setting discards the limitation that the event arguments exist within several consecutive sentences.
Arguments scattering throughout the long document would be in line with real-world applications and present more challenges for an event extraction model.
We set a baseline performance using \textsc{RolePred} on this dataset and provide insights for future work.

%% file: section/5_related.tex
\section{Related work}
\paragraph{Event Ontology Construction}  
Event ontologies are a crucial prerequisite to event discovery and extraction. 
Great efforts have been paid in previous studies to build several high-quality ontologies, such as FrameNet~\cite{baker1998berkeley},  Propbank~\cite{kingsbury2003propbank}, and VerbNet \cite{kipper2008large}.
However, it is costly and time-consuming to build hand-crafted ontologies. 
Some researchers start to explore automatic ontology construction.
Specifically, much progress has been made in event schema induction to characterize the relationship among different events \cite{cheung2013probabilistic, peng2016two, li2020connecting, kwon2020modeling, li2021future}.
Also, several recent studies attempt to discover new event types from raw texts \cite{shen2021corpus, edwards2022semi}.
Nevertheless, as the center of event ontologies, argument role prediction has always been an underexplored task. 
Related studies \cite{yuan2018open, liu2019open} restrict role names to a limited vocabulary so that they fail to reflect the unique characteristics of different event types.  
Therefore, in this paper, we study an essential but challenging task: open-vocabulary argument role prediction.

\input{table/template}

\paragraph{Event Extraction}
This task has been mainly studied under two paradigms:
(1) Sentence-level event extraction \cite{doddington2004automatic} has been studied since an early stage \cite{chen2015event, nguyen2016joint, yang2018dcfee}, with
a few models gone beyond individual sentences to make decisions \cite{ji2008refining, liao2010using, zhao-etal-2018-document}; and 
(2) document-level event extraction has gained a lot of research attention recently \cite{sundheim1992overview, du2020document, huang2021exploring, ma2022prompt, yang-etal-2021-document}. 
This study further explores extracting arguments scattered throughout documents.

%% file: table/template.tex
\renewcommand\arraystretch{1.0}
\begin{table*}[t]
\center \footnotesize
\tabcolsep0.10 in
\begin{tabular}{lcccccc}
\toprule

\textbf{Entity Type} & \textbf{Prompt} \\

\midrule

PERSON &  \textit{According to this, \underline{Entity} play the role of \lb MASK SPAN\rb in this \underline{Event Type}.}  \\
LOCATION &  \textit{According to this, the \lb MASK SPAN\rb is \underline{Entity} in this \underline{Event Type}.}   \\
NUMBER &  \textit{According to this, the number of \lb MASK SPAN\rb of this \underline{Event Type} is \underline{Entity}.}  \\
OTHER TYPES &  \textit{According to this, the \lb MASK SPAN\rb of this \underline{Event Type} is \underline{Entity}.}   \\

\bottomrule

\end{tabular}
\caption{Prompt design for different types of entities.}
\label{tab:template}
\vspace{-2mm}
\end{table*}

%% file: section/2_method.tex
\section{Method}

\begin{figure*}[t]
    \centering
    \includegraphics[width=0.97\linewidth]{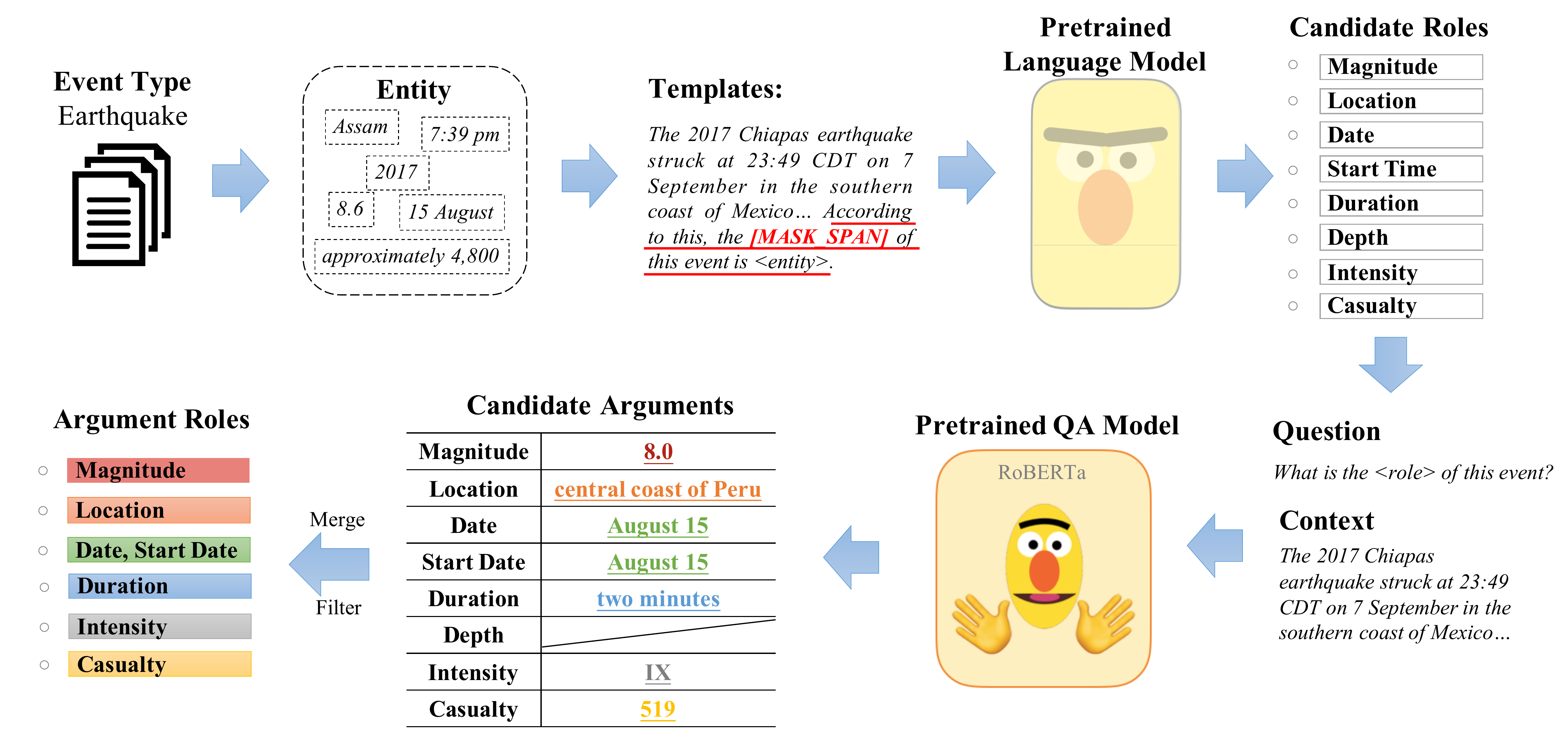}
    \caption{The framework of \textsc{RolePred}. It predicts argument roles by three components: first predict candidate role names for named entities by casting this problem as a prompt-based in-filling task, then extract candidate arguments for each candidate roles, and finally select the event-specific roles to serve for event extraction. }
    \label{fig:model}
\end{figure*}

\textsc{RolePred} contains three core components: candidate role generation, candidate argument extraction, and argument role selection (in Figure \ref{fig:model}). 
The following formulates the task of argument role prediction and then describes each component in turn.

\subsection{Task Formulation}
Formally, given an event type and a set of documents $\mathcal{D}$, each document $d \in \mathcal{D}$ mainly describes one event instance $e$ of the same type. The task of argument role prediction aims to predict a set of event-specific roles $\mathcal{R}$. Each role $r \in \mathcal{R}$ is a phrase or a cluster of phrases with similar semantics. 

\subsection{Candidate Role Generation}

Entities are often actors or participants in events. 
Thus, in the absence of available arguments, we introduce named entities to generate some candidates for argument roles.
Specifically, given an event type, for each document $d$, we use the off-the-shelf named entity recognition tool \cite{honnibal2017spacy} to identify all entities, $\mathcal{A}$, from the text.
Then, we treat these entities as possible arguments, and try to predict their roles. 
This process of candidate role generation is formulated as a mask-filling task. 
For each entity $a \in \mathcal{A}$, we construct a prompt with masked words to feed into the pre-trained language model. 
As a result, the model infers these masks as the role name of this entity by decoding its inner semantic knowledge.
Such a prompt is constructed as follows:

\textit{\underline{Context}. According to this, the \lb MASK SPAN\rb of this \underline{Event Type} is \underline{Entity}.}

Here \textit{Context} refers to the paragraph which mentions the entity from the source document. 
It provides a detailed background description of the event and the entity. 
Note that to avoid misleading information, the irrelevant sentences after the entity are removed. 
Then, it is followed by a natural language template containing \textit{\lb Entity\rb} and \textit{\lb Event Type\rb} placeholders. 
During inference, these placeholders are replaced by the concrete event type and entity.
\textit{\lb MASK SPAN\rb} represents a span of masks whose length is variable. 
For example, given the event type of earthquake and the entity of 5:36 PM, the constructed prompt is as follows:

\textit{The 1964 Alaskan earthquake, also known as the Great Alaskan earthquake, occurred at 5:36 PM AKST on Good Friday, March 27. According to this, the \lb MASK SPAN\rb of this \underline{earthquake} is \underline{5:36 PM}.}

In this case, \textit{\lb MASK SPAN\rb} is expected to be filled with \textit{time}, or \textit{start time} as the argument role.
Besides, considering the entity’s general semantic type: a person, location, number, or other, we slightly alter the prompt construction to fluently and naturally support the procedure of unmasking argument roles. Details are listed in Table \ref{tab:template}.

The constructed prompt is input into the encoder-decoder language model T5 \cite{raffel2020exploring} for candidate role generation. 
The generation process models the conditional probability of selecting a new token given the previous tokens and the input to the encoder. 
Note that the length of \textit{\lb MASK SPAN\rb} is not fixed for model filling. 
Inspired by SpanBERT \cite{joshi2020spanbert}, T5 samples the number of text spans from a Poisson distribution ($\lambda$ = 3). 
Each span is replaced with a single token. 
By infilling the marked text, the model is taught to predict how many tokens are missing from a span. 
Therefore, our roles generated by the language model are customized phrases of various lengths according to the semantics of constructed prompts
Unlike existing work that uses single general words as role names \cite{huang2016liberal, yuan2018open, liu2019open}, our roles are more fine-grained and contain more semantic details. 
This supports the subsequent task, argument extraction, to extract more participants for events from texts.
Finally, the language model generates 10 possible argument roles per entity.
For each document, we integrate the candidate role names of all entities for further selection.

\subsection{Candidate Argument Extraction}
For an event type, its salient argument roles are usually shared by most event instances. For example, each earthquake event has a magnitude but does not necessarily cause tsunamis. 
Therefore, it leaves the challenge of identifying relevant and salient roles for the candidates. 
Intuitively, arguments provide a feasible solution considering their strong interdependence with event roles.  
Along these lines, we first extract candidate arguments from each document for all candidate roles, and then conduct role selection using these arguments (more details in the next section). 

Inspired by some existing work on argument extraction \cite{lyu2021zero}, we formulate this problem into a question-answering task. 
Given an event type and a candidate role, we construct a question, which is fed into a standard pre-trained bidirectional transformer (BERT \citet{devlin2018bert}, RoBERTa \citet{liu2019roberta}) along with a document. 
The QA model serves to identify candidate event arguments (spans of text) from each source document.
Regarding the input sequences, we follow a standard BERT-style format as follows:

\textit{[CLS] What is the \underline{Event Role} in this \underline{Event Type} event? [SEP] \underline{Document} [SEP]}

Here, \textit{[CLS]} is BERT’s special classification token, \textit{[SEP]} is the special token to denote separation, and \textit{Document} is the tokenized input document.
For example, given the event type of pandemic, the event role of casualty, and a document on COVID-19, the input sequences are as follows:

\textit{[CLS] What is the casualty in this pandemic event? [SEP] The COVID-19 pandemic is an ongoing global pandemic of coronavirus disease. It's estimated that the worldwide total number of deaths has exceeded five million  ... [SEP]}

In this case, the argument is expected to be \textit{five million}. 
Note that, for some roles, a given document may not mention its argument. 
That is, the above-constructed question can be unanswerable.
Thus, for each extracted answer, we set a threshold on its probability from the QA model to filter out some unreliable results.
Besides, because our dataset focuses on one main event per document, unlike related work on sentence-level event extraction \cite{huang2020semi, liu2020event, ma2022prompt}, we need to search for arguments throughout the document. This task is more challenging and well worth further exploration.

So far, in every document, for each candidate role, one candidate argument has been extracted. 
Thus, these argument-role pairs can be composed into one event instance per document. 

\subsection{Argument Role Selection}

\begin{figure*}[t]
    \centering
    \includegraphics[width=0.97\linewidth]{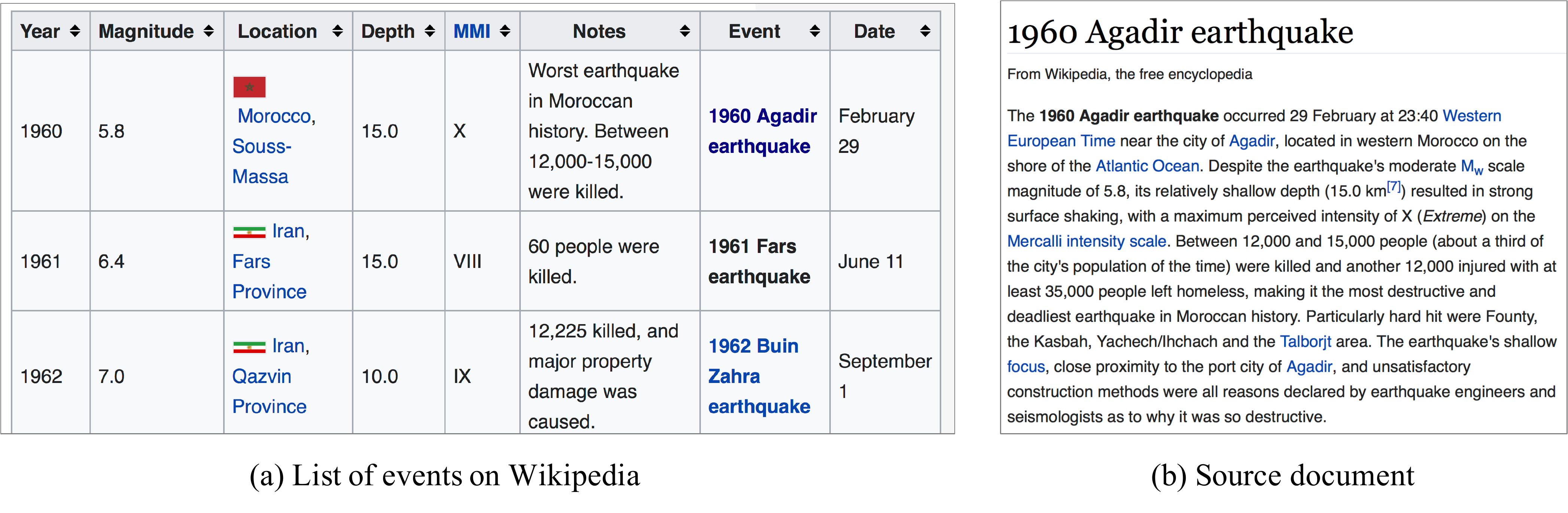}
    \caption{Data source of RoleEE. The left is a list of events from Wikipedia, from which we collect the argument roles and event instances on one event type. Each event instance has a URL pointing to its own Wiki page as shown in the right. We obtain the source documents from these Wiki page.}
    \label{fig:data}
    \vspace{-2mm}
\end{figure*}

After extracting the main event instance from each document, candidate roles are selected with mainly two steps: argument role filtering and merging.
Specifically, for an event type, its different event instances may present different attributes. 
These instances, however, usually have several common and significant argument roles (e.g., the intensity of the earthquake events and the host for the award ceremony events). 
Thus, we judge the salience of an argument role by involving multiple event instances of the same type.
It is assumed that a role name belongs to the event type only if most of the event instances have their associated argument. 

Regarding argument role merging, different roles can represent similar semantics and share the same arguments in an event. 
For example, the date, official date, and original date usually refer to the same day for a firework event. 
By merging similar role names, we can increase their specificity while reducing their number, thereby improving the efficiency of the subsequent argument extraction step. 
Along this line, we determine the semantic similarity of two roles based on the frequency that they share the same argument in the event instances.
For example, given 10 instances of the blizzard event, if two roles, data, and official date, have the same day as their arguments in 5 instances, then their similarity is 0.5.
We set a threshold to select semantically similar argument roles and merge them.

%% file: section/3_dataset.tex
\section{Dataset Construction}

\subsection{Data Collection}

\textbf{Event Type Selection.} Among the hot topics in the journalism, we carefully select 50 impactful event types, such as earthquake, civil unrest and military occupation. 
To broaden the domain coverage, these event types cover many fields including politics, academia, art, sport, military, astronomy, and economics.
Since these events usually contain rich argument roles, they require multiple sentences to describe. 
Thus, it is more suitable for document-level event extraction. 
More detailed examples are shown in Appendix \ref{sec:ap_data}. 

\noindent
\textbf{Argument Role Design.} To construct the event-specific argument roles, we leverage the list of events in Wikipedia.
Such a list shows the key attributes of multiple event instances of the same type. 
For example, Figure \ref{fig:data} show that the Wikipedia presents a list of recent major earthquakes.
Their attributes can be regarded as the prototype arguments of the event type, such as \texttt{Year}, \texttt{Magnitude}, \texttt{Location} and \texttt{Depth}.
Based on this observation, we search for one wiki list for each event type, and use the attributes as our basic set of argument roles.
Then, we manually process these argument roles:
(1) change abbreviations to common full names, such as \texttt{MMI} to \texttt{Magnitude}, (2) turn event names to triggers (\texttt{Name} or \texttt{Event} in the Wikipedia lists usually refer to the names of the event instances, which can be regarded as triggers), and
(3) remove \texttt{Notes} which adds extra details to the event instances, but not suitable to be an argument role. 
With such manual annotation from Wikipedia, we design customized argument roles for each event type.

\noindent
\textbf{Event Argument Annotation.} For each event type, the Wikipedia list usually involves multiple event instances. 
Each row in the list presents the information about one event.
The values of each row can be regarded as the arguments of an event.
For example, for the event ``1960 Agadir earthquake'', its magnitude is 5.8.
Further cleaning is conducted on event instances to ensure quality:  The event instances with incomplete arguments (e.g., null values or obvious errors) are dropped and
the event instances whose source documents are inaccessible are removed (document acquisition is introduced in the next section).
For the qualified events, their arguments are carefully refined by hand:
(1) save only the arguments of the selected roles,
(2) remove the special symbols or references in the arguments and keep only the key information, and
(3) discard the arguments which are not mentioned in the corresponding documents (since they may come from other sources and cannot be extracted from our documents).
Finally, for each event type, we obtain multiple event instances.

\noindent
\textbf{Source Document Acquisition.} 
For each event instance, we adopt its Wikipedia article as the source document where the event arguments are annotated. 
Specifically, the Wikipedia lists usually mention the event name and provide the URL of the corresponding Wikipedia article.
For example, as shown in Figure \ref{fig:data}, the first earthquake event is linked to the Wikipedia article of 1960 Agadir earthquake. 
These documents describe one major event and usually mention most of the event arguments in the Wikipedia lists. 
Otherwise, those arguments will be cleaned up.
We ensure that each event instance has a source document.
Besides, the documents with less than 4 sentences are removed.

\subsection{Data Analysis}
\input{table/dataset}

In total, RoleEE contains 50 event types and 142 unique argument roles. 
Each event type has 5.2 argument roles on average. 
It labels 4,132 valid document-level events and 15,562 event arguments. 
The event type of Championship has the highest average number of event arguments per document (8.5). 
We compare RoleEE to various representative event extraction datasets in Table \ref{tab:dataset}, including sentence-level EE datasets ACE2005 and KBP, and document-level EE dataset MUC-4, Wikievents, and RAMS. 
We find that RoleEE shows an advantage of rich argument role types, more than existing datasets.  
Besides some common argument roles, there are many unique role names customized for each event type.
Thus, our dataset is more versatile in this aspect.
In addition, RoleEE increases the difficulty in argument scattering, which is the critical challenge of document-level event extraction.
We count the number of sentences in which event arguments of the same event are scattered. 
As shown in Table \ref{tab:dataset}, the sentence-level EE event datasets only focus on one sentence, whereas our arguments are the most widely scattered among the document-level EE datasets, averaged with 7 sentences.
It calls for subsequent work to pay more attention to this challenge.

%% file: table/dataset.tex
\renewcommand\arraystretch{1.0}
\begin{table}[t]
\center \footnotesize
\tabcolsep0.05 in
\begin{tabular}{lcccccc}
\toprule

\textbf{Datasets} & \textbf{\# EvTyp.} & \textbf{\# RoleTyp.} & \textbf{\# Doc.} & \textbf{\# ArgScat.} \\

\midrule

ACE2005  & 33 & 35 & 599 & 1 \\
KBP2016  & 18 & 20 & 169 & 1 \\
KBP2017  & 18 & 20 & 167 & 1 \\
MUC-4    &  4 & 5 & 1,700 & 4.0 \\
WikiEvents  &  50 & 59 & 246 & 2.2 \\
RAMS     & 139 &  65 & 3,993 & 4.8 \\
\midrule
RoleEE   & 50 & 142 & 4,132 & 7.1 \\

\bottomrule
\end{tabular}
\caption{Statistics of EE datasets. \# EvTyp.: the number of event types. \# RoleTyp.: the number of unique argument roles. \# Doc.: the total number of documents. \# ArgScat.: the number of sentences in which event arguments of the same event are scattered. }
\label{tab:dataset}
\vspace{-2mm}
\end{table}

%% file: section/4_exp.tex
\section{Experiment}
\input{table/RNP_score}
In this section, we first study the performance on the argument role prediction task, then examine the performance on the downstream task, argument extraction. Finally, we report our case analysis.

\subsection{Argument Role Prediction}
\paragraph{Implementation}
Details about our implementation are introduced in Appendix \ref{sec:imple}. 

\paragraph{Baselines}
Our method is compared with four existing baselines:  LiberalEE \cite{huang2016liberal}, VASE \cite{yuan2018open}, ODEE \cite{liu2019open} and CLEVE \cite{wang2021cleve} (More information in Appendix \ref{sec:ap_baseline}). 
For ablation study, we evaluate three variants of \textsc{RolePred}: 
(1) - RoleMerge: it removes the similar role merging component from \textsc{RolePred} but still uses candidate arguments to filter those uncritical candidate roles;
(2) - RoleMerge - RoleFilter: it removes two components from the full model, including similar role merging and unimportant role filtering, which are introduced in Section 2.4;
and (3) \textsc{RolePred} (BERT): it adopts the same architecture of the full model while using the base version of BERT \cite{kenton2019bert} to generate candidate argument roles as introduced in Section 2.2. 
As to our full model, the base version of T5 \cite{raffel2020exploring} is utilized for candidate role generation.
In addition, we evaluate the human performance by inviting 3 PhD students who are not the authors of this paper to conduct this task manually.  
For each event type, each student is given the type name and 20 randomly selected documents. 
Then, each assessor writes down less than 20 argument roles, which are of less than three words. 
We average the scores of 3 students as the final human performance.

\paragraph{Evaluation Metrics} 
Following previous studies \cite{liu2019open}, we use precision, recall and F1-score as the metrics for argument role prediction. 
Two kinds of matching strategies are adopted: hard matching and soft matching.
The former requires that the generated argument role and groundtruth should have at least one word in common; whereas the latter
aims to compute the semantic similarity of a pair of roles.
Specifically, given two roles, we use a pre-trained bidirectional transformer, SentenceBERT \cite{reimers2019sentence}, to obtain their embeddings, and then calculate their cosine similarity as the semantic similarity score. 
Note that for multiple roles that are merged together, we concatenate them as one phrase for evaluation.

\paragraph{Experimental Result}
The evaluation results are shown in Table \ref{tab:rnp_score}, with the following observations. 
(1) Compared with the existing methods, \textsc{RolePred} achieves significant improvements of 18.4\% and 10.6\% on F1 scores with hard and soft matching respectively.
We speculate because argument roles are from open vocabulary.
This speculation is verified by checking other baselines.
LiberalEE and CLEVE perform relatively well since
their roles come from hand-crafted knowledge bases.
ODEE limits the roles to eight words, and the results validate its weaknesses.
VASE can't get roles explicitly, thus failing in the comparison.
(2) In the ablation study, even removing the merging and filtering parts, the variant of our method still outperforms the baselines, especially on hard matching. 
Based on this, role filtering provides a 4.0\% and 1.9\% improvement on F1 scores. 
The clear improvement of 2.7\% and 1.2\% occurs when we further merge similar roles.
As a base model, T5 generates better roles than BERT.
(3) The recall scores of \textsc{RolePred} can reach 64\% and 70\% on hard and soft matching respectively.
This indicates that the generated argument roles can cover most of the groundtruth. 
Likely, it benefits from a lot of diverse roles which involve various aspects of event types.
However, due to a large number of roles, the precision scores are reduced.
It suggests we carefully select important and relevant roles to ensure the efficiency of event extraction.

\subsection{Downstream Task}
We conduct experiments to investigate the effect of argument role prediction on its downstream task: argument extraction. 
This task aims to identify arguments directly from raw texts without available roles of the given event type.

\input{table/AE_wo_role_score}

\paragraph{Evaluation Metrics} 
We report precision (P), recall (R) and F1 scores as evaluation results. 
Note that the event arguments may have multiple mentions.
For example, the location of a fire can be a country, state, or city.
Therefore, we only require the extracted arguments to partially match with the groundtruth.
In addition, for those arguments of the date or time type, we normalize\footnote{\url{https://dateutil.readthedocs.io/en/stable/parser.html}} them into a uniform format for reasonable evaluation.

\paragraph{Baselines} 
LiberalEE, VASE, ODEE and CLEVE are our baselines, the same as the setting of argument role prediction.
Considering these methods extract multiple events from each document, we evaluate each with the groundtruth and then choose the highest score.
Besides, we also study three variants of \textsc{RolePred} for ablation study:

For ablation study, we evaluate three variants of \textsc{RolePred}: 
(1) - RoleMerge: it removes the similar role merging component from \textsc{RolePred} but still uses candidate arguments to filter those uncritical candidate roles;
(2) - RoleMerge - RoleFilter: it removes two components from the full model, including similar role merging and unimportant role filtering, which are introduced in Section 2.4;
and (3) \textsc{RolePred} (BERT): it adopts the same architecture of the full model while using the base version of BERT \cite{kenton2019bert} to extract candidate arguments as introduced in Section 2.3. 
As to our full model, the base version of Roberta \cite{DBLP:journals/corr/abs-1907-11692} is utilized for candidate argument extraction.
In addition, to explore the effect of role quality on downstream tasks, \textsc{RolePred}(Gold Roles) predict arguments using the true roles.  

\begin{figure}[t]
    \centering
    \includegraphics[width=0.9\linewidth]{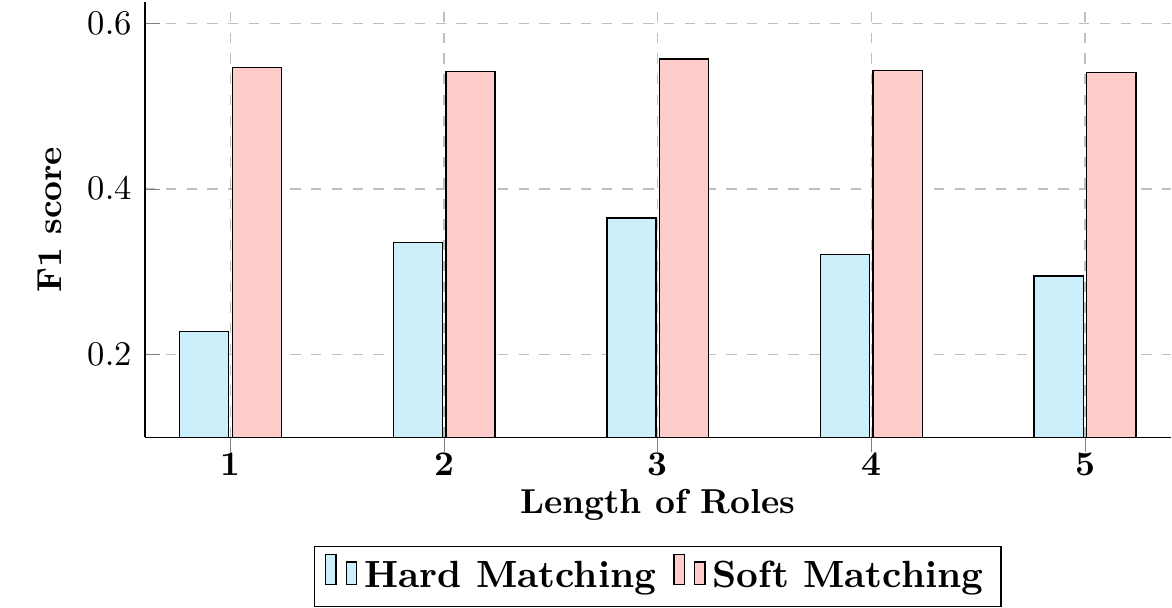}
    \caption{Impact of different length of role generation.}
    \label{fig:role_length}

\end{figure}

\paragraph{Experimental Result} \textsc{RolePred} and all its variants outperform other baselines by a large margin, as shown in Table \ref{tab:ae_wo_role_score}, 
This is likely because more specific role names can provide more semantics, thus assisting the model in identifying the correct arguments.
When comparing \textsc{RolePred} and its variants, a similar trend is observed under the evaluation setting.
By selecting salient roles, \textsc{RolePred} improves the effectiveness of argument extraction and increases the F1 score by 0.8\%.
Besides, by comparing with \textsc{RolePred} (Gold Roles), we can find that gold roles can greatly improve the precision score of argument extraction. 
Due to the large number of role names generated by the model, \textsc{RolePred} extracts more arguments and achieves a higher recall. 
Overall, given the high-quality roles, the f1 score of argument extraction is improved by 8\%.

\subsection{Impact of Role Length}
To explore the effect of role length on our task, we set different maximum lengths for candidate role generation.
Here we study the changing trend of f1 score using hard and soft matching.
According to Figure \ref{fig:role_length}, as the length increases from 1 to 5, the hard matching score shows a trend of increasing first and then decreasing, reaching a peak when the length is 3.
This shows that long roles can be somewhat fine-grained, but too much detail will introduce noises.
In addition, soft matching is not sensitive to this parameter.
We speculate that because the short role already covers key elements.

\subsection{Case Study}

\begin{figure}[t]
    \centering
    \includegraphics[width=1.0\linewidth]{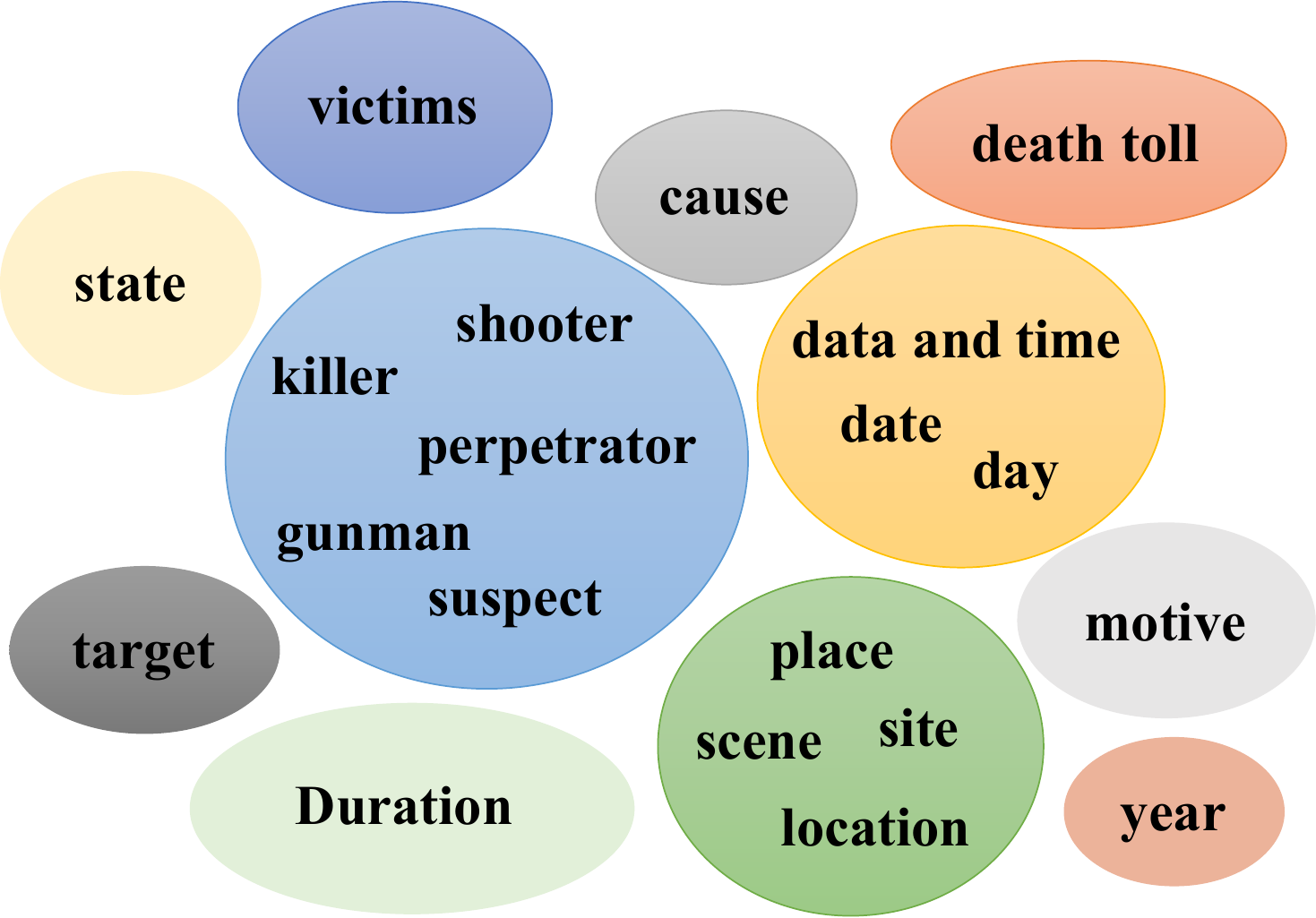}
    \caption{An example of our generated roles. The event type is \texttt{Shooting}. Each cluster has similar roles.}
    \label{fig:role_case}
    \vspace{-2mm}
\end{figure}

\begin{figure}[t]
    \centering
    \includegraphics[width=\linewidth]{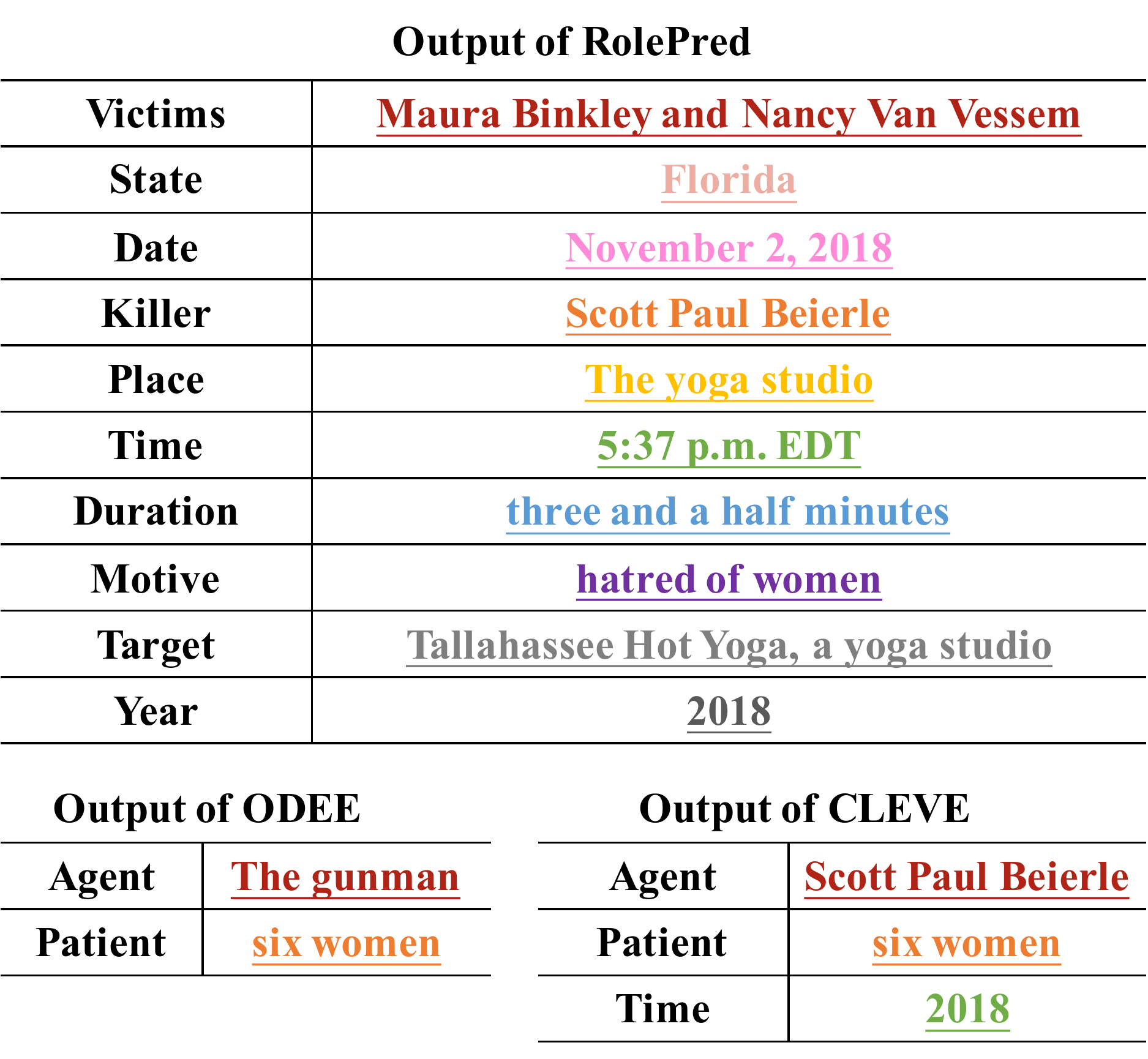}
    \caption{An example of the extracted events by different methods including ODEE, CLEVE, and \textsc{RolePred}. The event type is \texttt{Shooting} and the event instance is 2018 Tallahassee Shooting.}
    \label{fig:arg_case}
    \vspace{-2mm}
\end{figure}

An example of our generated roles is displayed in Figure \ref{fig:role_case}. 
The event type is \texttt{Shooting}.
Here, the roles with similar semantics are merged into the same cluster, such as killer, shooter, and suspect.
We can see that each cluster has various and salient roles for the shooting event. 
In addition, we also show a comparison of our model with baselines in Figure \ref{fig:arg_case} for the argument extraction task (one representative role is picked from the clusters). 
Benefit from rich roles, our model is able to effectively capture all arguments, while ODEE and CLEVE struggle with rare role types and result in uninformative extraction.
More cases can be found in Appendix \ref{sec:ap_case}. 

\subsection{Discussion on Data Leakage}
Since our argument extractor relies on RoBERTa trained on SQuAD v2.0 dataset, which comes from the same source of our constructed dataset RoleEE, it might lead to the data leakage risk. 
Thus, we exclude articles used in SQuAD v2.0 from RoleEE when constructing the dataset. 
Specifically, we compare all articles in our dataset with SQuAD2.0, and count the number of articles that share sentences with SQuAD2.0. 
Here we only consider sentences of more than 4 words. 
As the result, we remove all the overlapping articles from RoleEE. 
In this paper, the dataset statistics and the experiment results are reported after this process.

%% file: table/RNP_score.tex
\renewcommand\arraystretch{1.0}
\begin{table*}[t]
\center \footnotesize
\tabcolsep0.18 in
\begin{tabular}{lcccccc}
\toprule
\multicolumn{1}{l}{\multirow{2}[1]{*}{\textbf{Models}}} & \multicolumn{3}{c}{\textbf{Hard Matching}}
 & \multicolumn{3}{c}{\textbf{Soft Matching}} \\
 & \multicolumn{1}{c}{Precision} & \multicolumn{1}{c}{Recall} & \multicolumn{1}{c}{F1} & \multicolumn{1}{c}{Precision} & \multicolumn{1}{c}{Recall} & \multicolumn{1}{c}{F1}  \\
 
\midrule
LiberalEE & 0.1342 & 0.2613 & 0.1773 & 0.3474 & 0.5340 & 0.4209 \\
VASE & 0.0926 & 0.1436 & 0.1125 & 0.2581 & 0.4274 & 0.3218 \\
ODEE & 0.1241 & 0.3076 & 0.1768 & 0.3204 & 0.4862 & 0.3862 \\
CLEVE & 0.1363 & 0.2716 & 0.1815 & 0.3599 & 0.5712 & 0.4415 \\
\midrule
\textsc{RolePred} (BERT) & 0.2128 & 0.4582 & 0.2906 & 0.4188 & 0.6896 & 0.5211 \\
\textsc{RolePred} (T5) & \textbf{0.2552} & \textbf{0.6461} & \textbf{0.3659} & \textbf{0.4591} & \textbf{0.7079} & \textbf{0.5570} \\
\quad - RoleMerge & 0.2233 & 0.6962 & 0.3381 & 0.4234 & 0.7677 & 0.5457 \\
\quad - RoleMerge - RoleFilter & 0.1928 & 0.6582 & 0.2983 & 0.4188 & 0.7084 & 0.5264 \\
\midrule
Human & 0.6098 & 0.8270 & 0.7020 & 0.7365 & 0.8732 & 0.7990 \\

\bottomrule
\end{tabular}
\caption{Results of argument role prediction on our benchmark. Besides comparing with baselines, we also conduct the ablation study: the role merging and filtering are removed to verify their effectiveness.}
\label{tab:rnp_score}
\end{table*}

%% file: table/AE_wo_role_score.tex
\renewcommand\arraystretch{1.0}
\begin{table}[t]
\center \footnotesize
\tabcolsep0.08 in
\begin{tabular}{lcccccc}
\toprule

\textbf{Models} & \textbf{P} & \textbf{R} & \textbf{F1} \\

\midrule
LiberalEE & 0.2009 & 0.2941 & 0.2387 \\
VASE & 0.2123 & 0.3257 & 0.2570 \\
ODEE & 0.2402 & 0.3712 & 0.2917 \\
CLEVE & 0.3529 & 0.3890 & 0.3701 \\
\midrule 
\textsc{RolePred} (BERT) & 0.4170 & 0.4333 & 0.4250 \\
\textsc{RolePred} (Roberta) & \textbf{0.4131} & \textbf{0.5774} & \textbf{0.4817} \\
\quad - RoleMerge & 0.3855 & 0.6187 & 0.4750  \\
\quad - RoleMerge - RoleFilter & 0.4397 & 0.5001 & 0.4679 \\
\midrule 
\textsc{RolePred} (Gold Roles) & 0.6664 & 0.4948 & 0.5679 \\

\bottomrule
\end{tabular}
\caption{Results of argument extraction w/o gold roles. Besides the baselines, the argument merging and filtering are removed for ablation study.}
\label{tab:ae_wo_role_score}
\end{table}

%% file: section/6_conclusion.tex
\section{Conclusion}
This paper studies a challenging but essential task: open-vocabulary argument role prediction, and propose a novel unsupervised framework \textsc{RolePred} as a strong baseline and a carefully designed event extraction dataset for future work.

%% file: section/7_appendix.tex
\appendix

\begin{figure}[t]
    \centering
    \includegraphics[width=1.0\linewidth]{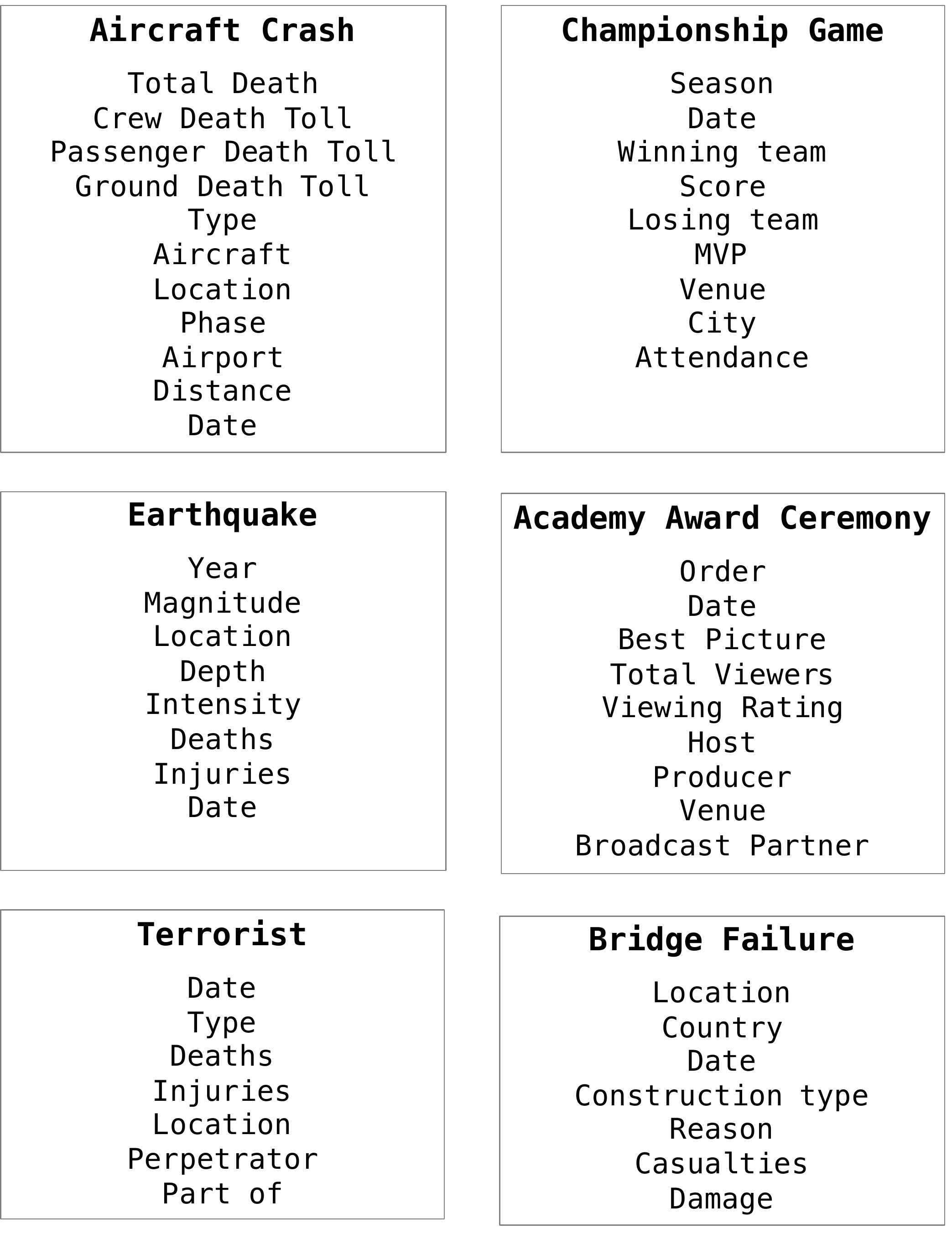}
    \caption{Examples of argument roles in our dataset.}
    \label{fig:example_role}
\end{figure}

\section{Dataset} \label{sec:ap_data}
\begin{figure*}[t]
    \centering
    \includegraphics[width=1.0\linewidth]{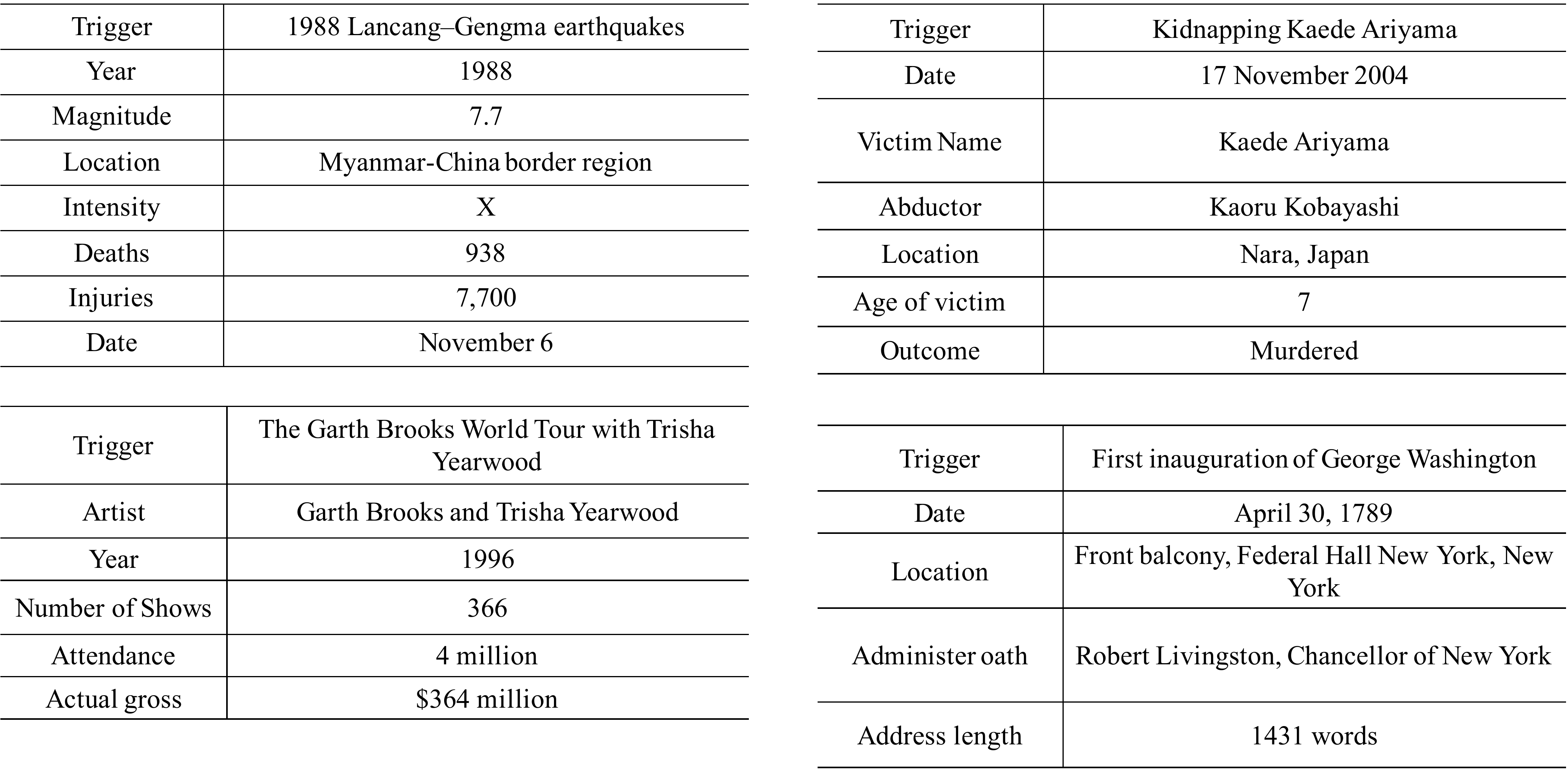}
    \caption{Examples of event arguments in our dataset.}
    \label{fig:example_arg}
\end{figure*}

In this section, we present more details about our dataset. 
All event types and the number of their corresponding documents are listed in Table \ref{tab:event_type}. 
In addition, we show some examples of argument roles of our dataset in Figure \ref{fig:example_role}.
Also, more examples of event instances are in Figure \ref{fig:example_arg}.

\section{Experiment} \label{sec:ap_exp}
\subsection{Implementation} \label{sec:imple}
To identify named entities from raw texts, we use the off-the-shelf named entity recognition tool from the SpaCy library\footnote{\url{https://spacy.io/}}.
For candidate role generation, we adopt the base version of T5 \cite{raffel2020exploring} as the pretrained generation model.
The model is built based on the Huggingface\footnote{\url{https://huggingface.co/models}}'s implementation with default parameters. 
The length of the constructed prompt is truncated to 512.
For each prompt, the model generates 10 sequences whose maximum length is 3.
The number of beams for beam search is set as 200. 
For candidate argument extraction, we use the large version of RoBERTa \cite{DBLP:journals/corr/abs-1907-11692} which has been trained on the SQuAD v2.0 benchmark \cite{rajpurkar-etal-2016-squad}. 
Its hyperparameters also refer to the Huggingface's implementation.
For the extracted argument, if its probability from the model is below 0.3, the argument is discarded.
For argument role filtering, given a role, when less than 40\% of the documents mention its corresponding argument, it will be filtered out.
For argument role merging, given a pair of roles, if they share the same argument in more than 50\% of the documents, they will be merged together.
We use one V100 GPUs with 32G memory for model training and evaluation. 
The prediction procedure lasts for about one day. 
For all the experiments, we report the average result of five runs as the final result. 
We also randomly select 20 documents for each event type and invite three students to annotate them for human evaluation. 

\subsection{Baselines}\label{sec:ap_baseline}
(1) LiberalEE \cite{huang2016liberal}: it leverages Abstract Meaning Representation to represent event structures and its argument roles are mapped with role descriptions in existing event knowledge bases \cite{baker1998berkeley, kingsbury2003propbank};
(2) VASE \cite{yuan2018open}: it proposes a Bayesian non-parametric model to obtain event profiles and represents argument roles with a tuple of entity roles; 
(3) ODEE \cite{liu2019open}: it constructs a latent variable neural model to extract unconstraint types of events from news clusters. and chooses argument roles from 8 possible reference words; and
(4) CLEVE \cite{wang2021cleve}: it provides a contrastive pre-training framework to learn event knowledge and follows the pipeline of LiberalEE to discover argument roles.

\begin{figure}[t]
    \centering
    \includegraphics[width=1.0\linewidth]{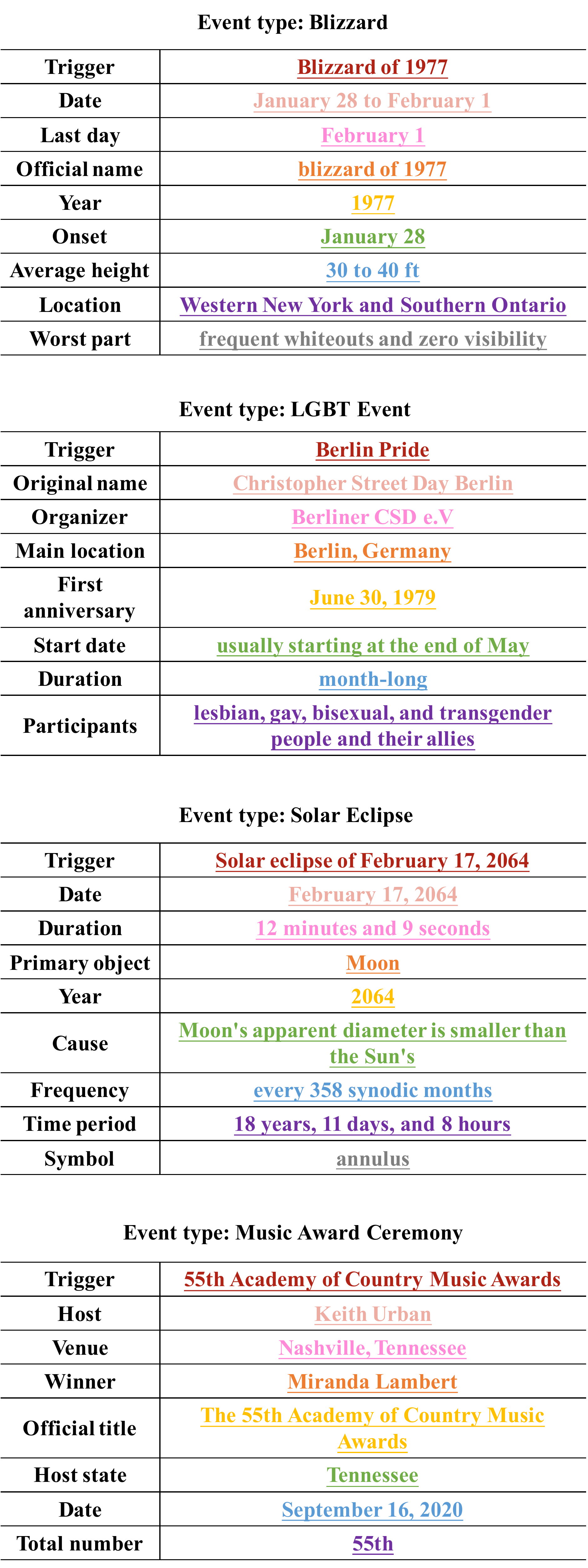}
    \caption{Four examples of event instances extracted by our framework.}
    \label{fig:more_case}
\end{figure}

\subsection{Case Study} \label{sec:ap_case}
\begin{figure*}[t]
    \centering
    \includegraphics[width=1.0\linewidth]{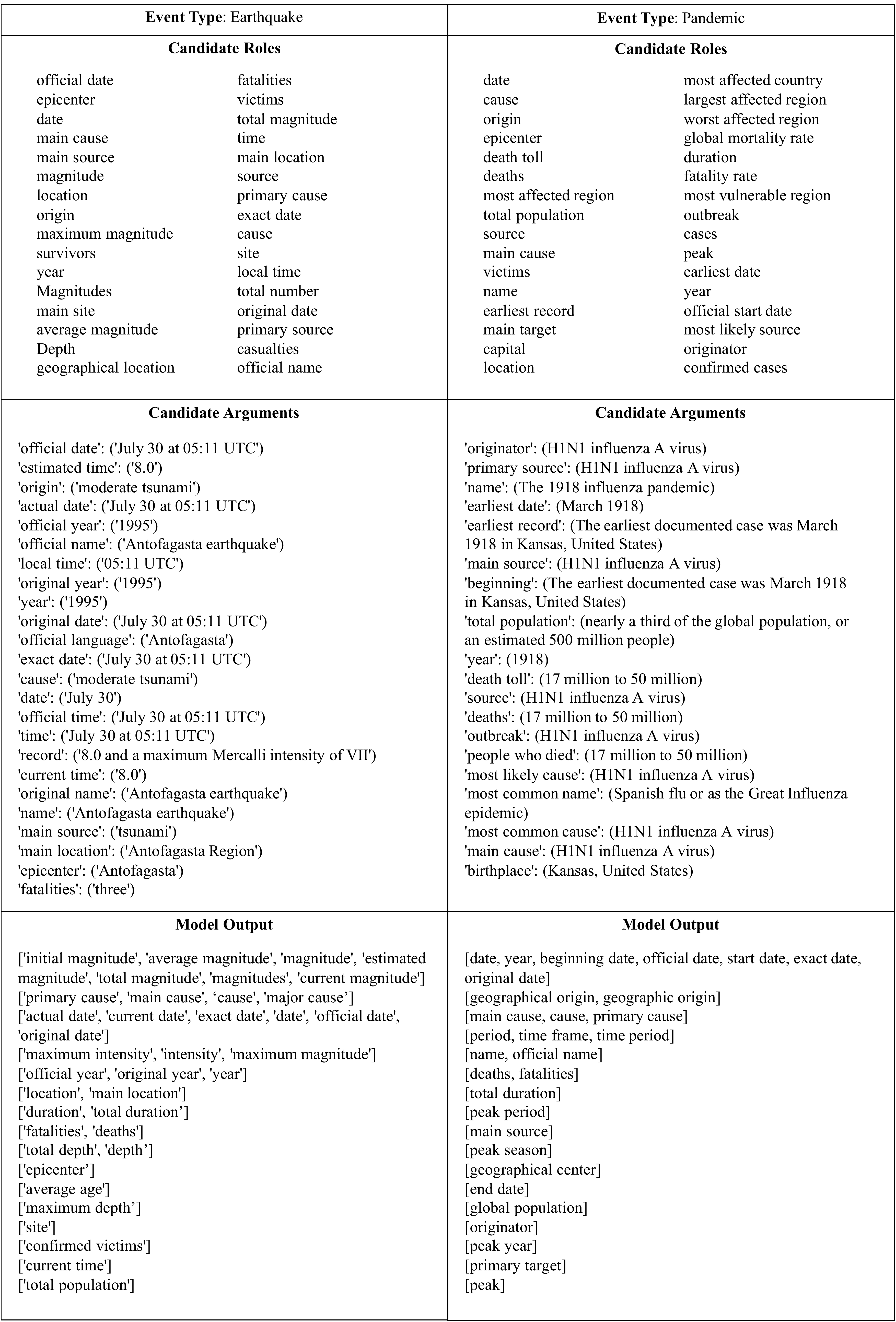}
    \caption{Example outputs of each component in our framework.}
    \label{fig:full_case}
\end{figure*}

To study each component in our framework, we show two examples of their outputs given two event types of Earthquake and Pandemic in Figure \ref{fig:full_case}. 
The example includes the generated candidate roles, the extracted candidate arguments, and the clusters of roles as the final model output. 
Here, the generated candidate roles are sorted by their importance scores.
The extracted candidate arguments are from a randomly selected document.
And the clusters of roles are ranked by the cluster size and the importance scores.
In addition, Figure \ref{fig:more_case} presents four more extracted event instances. 
We remove the roles that have no available argument in the source document. 
From these cases, we can see that our method can actually 
extract informative and reasonable events with specific argument roles.

\input{EMNLP 2022/table/event_type}

%% file: EMNLP 2022/table/event_type.tex
\renewcommand\arraystretch{1.2}
\begin{table*}[t]
\center \footnotesize
\tabcolsep0.12 in
\begin{tabular}{ccccccc}
\toprule

\textbf{Event Type} & \textbf{ \# Docs} & \textbf{Event Type} & \textbf{ \# Docs} & \textbf{Event Type} & \textbf{ \# Docs}  \\

\midrule

film festival & 532 & aviation accident & 459 & aircraft crash & 390 \\ 
massacre & 222 & kidnapping & 216 & explosion & 190 \\
flood & 178 & war & 147 & LGBT event & 130 \\
satellite launch & 130 & military occupation & 117 & bridge failure & 100 \\
shipwreck & 94 & disaster & 91 & sentence & 91 \\
human stampede & 84 & NBA final & 77 & concern tour & 71 \\
dam failure & 68 & inauguration & 63 & Olympic game & 62 \\
earthquake & 61 & tornado & 57 & railway terrorist & 49 \\
resignation & 47 & strike & 47 & academy award ceremony & 44 \\
avalanche & 43 & boiler explosion & 39 & blizzard & 33 \\
terrorist & 32 & firework & 32 & bank failure & 32 \\
civil unrest & 29 & extinction & 28 & music award ceremony & 28\\
wildfire & 26  & bushfire & 26 & bankruptcy & 23 \\
protest & 23 & surfing competition & 23  & shooting & 22 \\
pandemic & 21  & rail accident & 21 & volcano eruption & 21 \\
recession & 19 & solar eclipse & 17  & nightclub fire & 17\\
festival & 17 & championship game & 14 & & \\

\bottomrule
\end{tabular}
\caption{All the event types and the numbers of corresponding documents in our dataset.}
\label{tab:event_type}
\end{table*}